\documentclass[letterpaper]{article} 
\usepackage{aaai20}  
\usepackage{times}  
\usepackage{helvet} 
\usepackage{courier}  
\usepackage[hyphens]{url}  
\usepackage{graphicx} 
\usepackage{graphicx}  
\usepackage{latexsym}
\usepackage{hyperref}
\usepackage{hyperref}
\usepackage{flafter}

\usepackage{float}

\usepackage{amssymb}
\usepackage{amsmath}
\usepackage{bbm}
\usepackage{todonotes}
\usepackage{framed}
\usepackage{multirow}
\usepackage{array} 
\usepackage{bigints}
\usepackage{color}
\usepackage{soul}
\usepackage{xcolor}

\usepackage{tikz}
\usetikzlibrary{shapes.geometric}

\usepackage{lipsum}
\usepackage{lmodern}
\usepackage{tcolorbox}
\usepackage{placeins}

\usepackage[linesnumbered,ruled]{algorithm2e}
\usepackage{subfig}
\usepackage{graphics}
\usepackage{multirow}
\usepackage{multicol}

\usepackage{cleveref}
\usepackage{xcolor,colortbl}

\usepackage{scalerel}

\usepackage{caption}
\definecolor{green}{rgb}{0,1,0}
\definecolor{red}{rgb}{1,0,0}

\newcommand\score[2]{
\pgfmathsetmacro\pgfxa{#1+1}
\tikzstyle{scorestars}=[star, star points=5, star point ratio=2.25, draw,inner sep=1.3pt,anchor=outer point 3]
  \begin{tikzpicture}[baseline]
    \foreach \i in {1,...,#2} {
    \pgfmathparse{(\i<=#1?"yellow":"gray")}
    \edef\starcolor{\pgfmathresult}
    \draw (\i*1.75ex,0) node[name=star\i,scorestars,fill=\starcolor]  {};
   }
  \end{tikzpicture}
}

\makeatletter
\newcommand{\printfnsymbol}[1]{%
  \textsuperscript{\@fnsymbol{#1}}%
}
\newcommand{\mat}[1]{\mathbf{#1}}

\newcommand{\cC}{\mathcal{C}}

\DeclareMathOperator{\argminA}{arg\,max}

\def\thumbsup{\scalerel*{\includegraphics{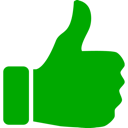}}{O}}
\def\thumbsdown{\scalerel*{\includegraphics{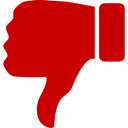}}{g}}

\makeatletter

\usepackage[switch]{lineno} 

\usepackage{microtype}

\title{An Embedding-based Joint Sentiment-Topic Model for Short Texts}

\author{
\Large \textbf{Ayan Sengupta\thanks{Equal Contribution.}\thanks{Corresponding Author.}$^{\rm 1}$\thanks{The paper has been accepted as a full paper in International AAAI Conference on Web and Social Media (ICWSM), 2021.}, William Scott Paka\printfnsymbol{1}\thanks{This work was done when the author was an intern at OGA during May-July 2019.}$^{\rm 2}$, Suman Roy$^{\rm 1}$, Gaurav Ranjan$^{\rm 1}$, Tanmoy Chakraborty$^{\rm 2}$ }\\
\textsuperscript{\rm 1}Optum Global Advantage (OGA),  (UnitedHealth Group)  India\\ 
\{ayan\_sengupta, suman.roy, gauravranjan\}@optum.com\\
\textsuperscript{\rm 2}Dept. of CSE, IIIT-Delhi, India\\
\{william18026, tanmoy\}@iiitd.ac.in\\
}
\date{}

\begin{document}
\maketitle

\begin{abstract}
Short text is a popular avenue of sharing feedback, opinions and reviews on social media, e-commerce platforms, etc. Many  companies need to extract meaningful information (which may include thematic content as well as semantic polarity) out of such short texts to understand users’ behaviour. However, obtaining high quality  sentiment-associated and human interpretable themes still remains a challenge for short texts. In this paper we develop ELJST, an embedding enhanced generative joint sentiment-topic model that can discover more coherent and diverse topics from short texts. It uses Markov Random Field Regularizer that can be seen as  generalisation of skip-gram based models. Further, it can leverage higher order semantic information appearing in word embedding, such as self-attention weights in graphical models. Our results show an average improvement of 10\% in topic coherence and 5\% in topic diversification over baselines. Finally, ELJST helps understand users' behaviour at more granular levels which can be explained. All these can bring significant values to service and healthcare industries often dealing with customers.

\end{abstract}

\section{Introduction}
Short text is a popular mean of communication in online social media and e-commerce websites that appear abundant in different applications. 
Mining short texts is thus essential to extract thematic content of the text as well as to identify the sentiment expressed by the customers about certain entities (products, services, and movies to name a few). 
In many applications it may be required to discover both topic and sentiment simultaneously as seen in target dependent (or topic-specific) sentiment analysis~\cite{gupta2019multi}. 

\noindent{\bf A Motivation for this work:}\,
There have been a few attempts to predict both sentiment and topics simultaneously~\cite{MeiSZ07,LinHER12,RW16,Nguyen2015TopicMB}; among which extraction of Joint Sentiment-Topic (JST) model is quite popular. Let us illustrate the functionality of JST compared to ELJST (our proposed method to be introduced subsequently) through the following example of a review:

\medskip
\noindent
\framebox{%
 \parbox[t][1.42cm]{7.35cm}{ {\it Claims} {\it payment} was fast and easy. However, {\it language barrier} with {\it customer} {\it care} was really {\it difficult} to deal with. \score{4}{5}
 }}
 \smallskip


\smallskip

\hspace*{-.5cm} JST: \\ \fbox{\small claims \thumbsup\, 0.8} \fbox{\small pay \thumbsup\, 0.8} \fbox{\small customer \thumbsdown\, 0.45}
\vspace*{+.16cm}\\
JST with skip-gram: \\ \fbox{\small claims pay\, \thumbsup\, 0.7} \fbox{\small customer care difficult \thumbsdown\, 0.6} \hspace*{-.5cm}
\vspace*{+.19cm}\\
ELJST:\\
\hfill\rlap{\fbox{\small claims pay fast \thumbsup\, 0.85} \fbox{\small customer language barrier \thumbsdown\, 0.7}}

\vspace*{+.2cm}

JST will discover the topics such as {\tt claims}, {\tt pay}, {\tt  customer}, etc. Also using  skip-gram-based JST model (a skip-gram JST can be developed by assuming a topic distribution over $n$-grams) one will be able to discover topics like {\tt claims pay}, {\tt customer care difficult} etc. Through the use of an appropriate sentiment lexicon, JST will also detect  sentiment values of the topics as shown above, without considering any external sentiment labels (star rating). However, JST suffers from few drawbacks such as using  only unlabeled data, for which it is unable to incorporate external labels like the ratings given by the customers or, ground-truth labels obtained from the annotators etc. We believe that external labels often play an important role in determining sentiment and topics jointly. For instance, in the above example, the 4-star rating given by the customer can be incorporated to better identify the sentiment of the topics. Also JST does not allow context-based information to be used for model discovery, which otherwise using skip-gram model, may lead to better topic quality as we will see later in this paper.

To alleviate these issues, we introduce {\bf Embedding enhanced Labeled Joint Sentiment Topic (ELJST) Model}, a novel framework that jointly discovers topics and sentiment for short texts in presence of labeled (with discrete sentiment values) texts. ELJST model bears close resemblance to the work of weakly supervised joint topic-sentiment model~\cite{LinHER12}, which is an extension of the classic topic model based on Latent Dirichlet Allocation (LDA)~\cite{BleiNJ03}. ELJST constructs an extra sentiment layer on top of LDA  with the assumption that sentiments are generated based on topic distributions, and words are generated by conditioning on the topic-sentiment pairs. To skirt the sparsity problem we do not take recourse to usual practice of topic modeling on short texts using skip-gram~\cite{ShiKCR18} or, bi-term~\cite{Yan+13} as these models are inherently required to be parameterized with the window size of the context or the length $n$ of n-grams. Rather we use Markov Random Field (MRF) Regularizer that creates an undirected graph for each text by constructing edges between contextually and semantically similar words, and formulates a well-defined potential function to enhance topic identification.

\medskip
\noindent
 \textbf{Contributions of our work:}
  \begin{itemize}
 \item Our model can generalise skip-gram or $n$-gram based joint topic-sentiment models by considering long-term dependency between tokens by leveraging embeddings or, self attentions.
 \item Further, by using overall text labels in a supervised manner, we can avoid using external lexicons and incorporate richer domain specific external knowledge into generative models.
 \item Our model produces tighter and more coherent latent representation by employing MRF regularizer, which makes topic models to be more human interpretable for short texts.
 \end{itemize}
 
 {\color{black}For reproducibility, we have uploaded the public datasets and our codes to \url{https://github.com/DSRnD/ELJST}}.
 

\section{Related Work}
We briefly describe prior art in two parts, that of related to --  (1) Joint Sentiment-topic Extraction, and (2) Word Embedding Assisted Topic Extraction and Sentiment Modeling.
Due to the abundance of literature on sentiment analysis and topic modeling, we restrict to studies which we deem pertinent to our work. 

\paragraph{\bf Joint Sentiment-Topic Extraction:}   Topic-sentiment Model (TSM)~\cite{MeiSZ07} is the first attempt to deal with the extraction of sentiment and topic models jointly.
As TSM is primarily based on probabilistic latent semantic indexing (pLSI)~\cite{Hofmann99}, it suffers from two common drawbacks: inferring quality topics for new document and over-fitting. To overcome these,~\cite{LinHER12} propose a weakly supervised hierarchical Bayesian model, {\em viz.}  JST. 
The extraction of JST model ensures topic generation to be conditioned on sentiment labels. The same authors introduce another model, called Reverse-JST (RJST) in which sentiment generation depends on topic.
Sentiment-LDA (sLDA)~\cite{Li10} and Dependency-Sentiment-LDA (dsLDA)~\cite{Li10} use external sentiment lexicon under global and local context to be linked with topic identification from texts. The labeled topic model proposed in~\cite{RamageHNM09} uses external labels for capturing linear projection on the priors. However, all these models are unable to discover fine-grained dependency between topics and sentiments. To address this, hidden topic-sentiment model (HTSM) is introduced that explicitly captures topic coherence and sentiment consistency from opinionated texts~\cite{RW16}. \cite{PoddarHL17} propose SURF that identifies opinions expressed in a review.
\cite{Nguyen2015TopicMB} introduce Topic Sentiment Latent Dirichlet Allocation (TSLDA), a new topic model that can capture the topic and sentiment simultaneously. \color{black}One of the popular topic models employed to automatically extract topical contents from the documents is based on  non-negative matrix factorization (NMF) {\em e.g.}, ~\cite{Lee99,XLG03}. However, this usually does not produce sentiment labels. For this, one has to address sentiment prediction problem for short texts (some of which are labeled with discrete or real numbers) using a semi-supervised approach of extracting joint sentiment/topic model. Such a method has been proposed in~\cite{LiZS09} using a constrained non-negative tri-factorization of the term-document matrix implemented using novel yet simple update rules; however, it uses discrete sentiment values only. The authors extended this approach to incorporate real sentiment values lying in a particular range~\cite{RAM18}.
But this cannot match up to the accurate sentiment values predicted by other methods for which we do not consider this work for benchmarking purpose. Guha et al. \cite{guha2016tweetgrep} proposed a weekly supervised approach which models topics and sentiments jointly.

\color{black}

\begin{table*}[ht]
\caption{
    Comparison of ELJST and other baseline methods w.r.t different dimensions (T: Topic, S: Sentiment, W: Word, Doc: Document, UL: Unlabeled, L: Labeled).}
\centering
\small
\scalebox{1}{
    \begin{tabular}{|c|c|c|c|c|c|c|}
    \hline
    \multirow{4}{*}{Model} &Input &	Lexicon  & Word embedding & \multicolumn{3}{c|}{Output}\\
    \cline{5-7}
    & data & needed? & used? & \multirow{2}{*}{T-S} & T dist.  &	T \\
    & & & & & over W & polarity\\
    \hline
    JST~\cite{LinHER12} & UL  &	Yes & No	& T under S  &	Global &	Doc-level\\
    
    TSM~\cite{MeiSZ07}	&  UL  &	Yes &	No & T-S pair  &	Local &	Global\\

    RJST~\cite{LinHER12} &	UL	& Yes & No &	T-S pair  &	Local &	Doc-level\\ 
    WS-TSWE~\cite{kbs/FuSWCH18} &	UL	& No & Yes &	T-S pair  &	Local &	Doc-level\\ \hline
    {\bf ELJST} & L  &	No &	Yes & T-S pair  &	Local &	Doc-level\\
    \hline
     \end{tabular} }
     \vspace{1mm}
    \label{tab:comparison}
\end{table*}

\paragraph{\bf Word Embedding Assisted Topic Extraction and Sentiment Modeling:}
Recently, researchers have started using richer word representations to fine-tune topic models in order to extracting more meaningful topics~\cite{QiangCWW17,kbs/FuSWCH18}. As mentioned in~\cite{QiangCWW17}, an embedding-assisted topic model can understand the latent semantic relationship between two words ``king" and ``queen" and place them under the same topic, irrespective of whether they co-occur in same text or not. Another advantage of using word embedding in topic models lies in its ability to generalize the model. The authors in~\cite{Yan+13} use bi-grams instead of unigram in order to tackle sparsity in short texts. 
Recently in \cite{kbs/FuSWCH18,ecml/FuSWCH16}, the authors propose a novel topic sentiment joint model called weakly supervised topic sentiment joint model with word embedding (WS-TSWE), which incorporates word embedding and HowNet lexicon simultaneously to improve the topic identification and sentiment recognition. 
A generalized model is introduced in~\cite{kbs/AliKKEAUKK19} which is able to use $n$-grams to capture long term dependencies between words. The work on joint Sentiment Topic model aims to deal with the problem about the mixture of topics and sentiment simultaneously. Most of them have gone to show that embedding and joint sentiment-topic joint model can be combined effectively to discover the mixture of topics and sentiment simultaneously.

Table~\ref{tab:comparison} summarizes a comparison of ELJST with existing models w.r.t. different dimensions of the model.

\section{Embedding enhanced Labeled Joint Topic-Sentiment Model}
In this section, we discuss the proposed Embedding enhanced Labeled Joint Topic-Sentiment model (ELJST) for identifying  coherent and diverse topics along with sentiment classes extracted from labeled text data. 


\subsection{Our Proposed Model} \label{proposedModel}
Let\, $\cC = \{d_1, d_2, \ldots, d_D\}$ denote a collection of $D$ documents. A document $d =w_1, w_2, \ldots, w_{N_d}$ is represented by a sequence of $N_d$ words. Distinct words are indexed in a vocabulary ${\cal V}$ of size $V$.  Also let $S$ and $T$ be the number of distinct sentiment labels and  topics respectively. 
We assume each document $d$ to be labeled with a number ${ \lambda}^{d} \in \{1,2, \ldots S\}$. This allows to define a document-specific label projection vector $\mat{L}^{(d)}$ of dimension $S$ as:
\color{black}
\[
 L_k^{(d)} = \begin{cases}
                1 &  \rm{if} \,\,\,\lambda^{d} $ = \textit{k}$        \\
                0 & \rm{otherwise}
                   \end{cases} 
                   \]
\\

In other words, the $k$th entry of ${\bf L}^d$ is $1$ if the label of document $d$ is $k$. 
\color{black}
We approximate it by $L^{(d)} \leftarrow L^{(d)} + \epsilon,\, 0< \epsilon < 1$. 
In ELJST model, we say two words are semantically similar if their distance is less than a threshold value. We create an undirected graph $G$ (MRF) for each document $d$ by connecting semantically similar words and their corresponding topic assignments. We identify semantic similarity between two words using appropriate distance metric on various embedding representations,  such as Word2Vec~\cite{MikolovSCCD13,corr/GoldbergL14}, sub-word level representation (fastText) ~\cite{BojanowskiGJM16,li-etal-2018-subword}, contextual embedding (BERT)~\cite{DevlinCLT19} and attention models ~\cite{BahdanauCB14,VaswaniSPUJGKP17}. Each of these techniques have their own merits and demerits. Word2Vec is easy to use, although, word embedding for domain specific words are not always available on Word2Vec. fastText~\cite{BojanowskiGJM16} uses sub-word level representations and can generate word vectors for out of vocabulary words. However, both Word2Vec and fastText produce static embeddings.
On the other hand, BERT~\cite{DevlinCLT19} can capture contextual information which allows one to construct dynamic edges for same word token in different contexts. 

\subsection{Generative Model of ELJST} \label{ELJST}
\begin{figure}[!t]
     \centering
     {\includegraphics[scale=0.45]{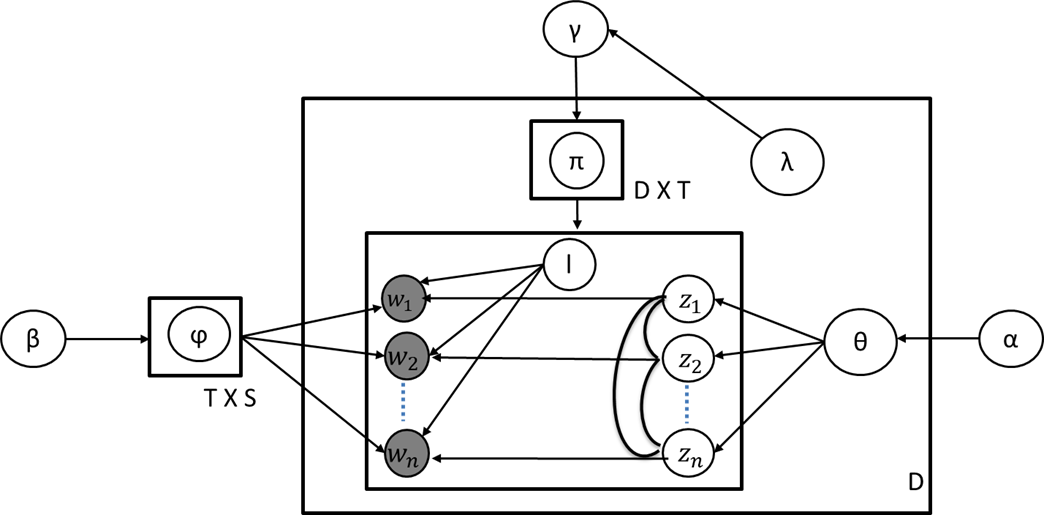}\label{ejst}}
     \caption{Generative Model of ELJST.}
     \label{figELJST}
\end{figure}

Generating a word $w_i$ in document $d$  is  a three-stage procedure, as shown in Figure~\ref{figELJST}. First, a topic $j$ is chosen from a per-document topic distribution ${\theta}_d$. Following this, a sentiment label $l$ is chosen from sentiment distribution ${\pi}_{d,j}$, which is conditioned on the sampled topic $j$. Finally, a word is drawn from the per-corpus word distribution conditioned on both topics and sentiment labels $\varphi_{j,l}$. The steps for the generative process in ELJST shown in Figure~\ref{figELJST} are formalised as below:

\FloatBarrier
\begin{table} [htb]
\begin{center}
\begin{tabular}{ll}
{\bf 1.} & For  each document $d$ \\
& \,\,\, Generate ${\theta}_d \sim {\rm Dir}(\pmb {\alpha})$;\\
{\bf 2.} & For each document $d$ and \\
&  \,\,\, topic $j \in \{1, 2, \ldots T\}$ \\
  &  \,\,\, Choose ${\pi}_{d,j}  \sim {\rm Dir}(\pmb{\gamma}^{(d)})$, ${\pmb{\gamma}}^{(d)} = \gamma \times \mat{L}^{(d)}$;\\
{\bf 3.} & For  each topic $j \in \{1, 2, \ldots T\}$ and \\ 
  & \,\,\, sentiment label $l \in \{1, 2, \ldots S\}$ \\
 &  \,\,\, Choose $\varphi_{j,l} \sim {\rm Dir}(\beta)$;\\
{\bf 4.} &  For  each word $w_i$ in document $d$\\
&  \hspace*{0.2cm} {\bf (a)} Choose topic $z_i \sim {\rm Mult}(\theta_d)$;\\
& \hspace*{0.2cm} {\bf (b)} Choose sentiment label \\
& \,\,\, $l_i \sim {\rm Mult}(\pi_{d,z_i})$;\\
&  \hspace*{0.2cm} {\bf (c)} Choose word $w_i \sim {\rm Mult}(\varphi_{z_i, l_i})$, \\
 & \,\,\, a multinomial distribution over words \\
& \hspace*{0.2cm} conditioned on sentiment label $l_i$ and \\
& \hspace*{0.2cm} topic $z_i$.\\
    \end{tabular} 
    \end{center}
    \end{table}

Here, $\pmb{\alpha}$ and $\beta$ are hyperparameters -- the former is the prior observation count, denoting the number of times topic $j$ is associated with document $d$, and the latter is the number of times words sampled from topic $j$ which are associated with sentiment label $l$  before observing the actual words. ${\rm Dir}(\cdot)$ is the Dirichlet distribution. The hyperparameter $\pmb{\gamma}$ indicates the prior observation number that counts how many times a document $d$ will have the label $l$ before any word from the document is observed. We also use the vector $\mat{L}^{(d)}$ to project the parameter vector of the Dirichlet document sentiment prior ${\pmb{\gamma}} = (\gamma_1, \gamma_2, \ldots, \gamma_{S})$, to a lower dimensional vector~\cite{RamageHNM09}:
\color{black}
\[ {\pmb{\gamma}^{(d)}} = {\mathbf \gamma} \times \mat{L}^{(d)} =
 \begin{cases}
                (1+\epsilon)\gamma &  \rm{if} \,\,\,\lambda^{\textit{d}} $ = \textit{k}$        \\
                \epsilon \gamma & \rm{otherwise}
                   \end{cases} 
                   \]
\\
\color{black}
The perturbation parameter $\epsilon$ is used to forcibly assign non-zero values to labels. We have used $\pmb{\alpha}$ and $\pmb{\gamma}$ as the asymmetric priors and $\beta$ as the symmetric prior. We need to infer three sets of latent variables -- per-document topic distribution $\theta$, per-document topic specific sentiment distribution $\pi$, and per-corpus joint topic-sentiment word 
distribution $\varphi$.
\subsection{Model Inference and Parameter Estimation}

The joint probability of the words, topics and sentiment labels can be decomposed as follows:
\begin{equation}\label{jointpWTSnew}
p(\mathbf{w}, \mathbf{z}, \mathbf{l}) =  p(\mathbf{w}\,|\, \mathbf{l}, \mathbf{z}) \cdot  p(\mathbf{l}, \mathbf{z})  =  p(\mathbf{w}\,|\, \mathbf{l}, \mathbf{z}) \cdot p(\mathbf{l}| \mathbf{z}) \cdot p(\mathbf{z}) 
\end{equation}

The first term of Eq.~\ref{jointpWTSnew} is obtained by integrating w.r.t. $\varphi$ shown in Eq.~\ref{probw}, where $N_{j,k,i}$ is the number of times word $i$ appears in topic $j$ with sentiment label $k$, and $N_{j,k}$ is the number of times words are assigned to topic $j$ with sentiment label $k$.

\begin{equation}\label{probw}
p(\mathbf{w}\,|\, \mathbf{l}, \mathbf{z}) = {\left(\frac{\Gamma(V\beta)}{{\Gamma(\beta)}^{V}} \right)}^{T \times S} \cdot \prod\limits_{j} \prod\limits_{k} \frac{\prod\limits_{i} \Gamma(N_{j,k,i}+\beta)}{\Gamma(N_{j,k}+V \beta)} 
\end{equation}

The second term  of Eq.~\ref{jointpWTSnew} is obtained by integrating w.r.t. $\pi$ shown in Eq.~\ref{probl}, where, $N_{d,j,k}$ is the number of times a word from document $d$ is associated with topic $j$ and sentiment label $k$, and $N_{d,j}$ is the number of times topic $j$ is assigned to some word tokens in document $d$.

\begin{equation} \label{probl}
p(\mathbf{l}| \mathbf{z}) = {\left(\frac{\Gamma(\sum\limits_{k=1}^{S}\gamma_{d,k})}{\prod\limits_{k=1}^{S}\Gamma(\gamma_{d,k})}\right)}^{D \times T} \cdot \prod\limits_{d} \prod\limits_{j} \frac{\prod\limits_{k} \Gamma(N_{d,j,k}+\gamma_{d,k})}{\Gamma(N_{d,j}+ \sum\limits_{k} \gamma_{d,k})}
\end{equation}

We write the third term  of Eq.~\ref{jointpWTSnew} by integrating w.r.t. $\theta$, as shown in Eq.~\ref{probz}, where $N_d$ is the total number of words in document $d$. As discussed in~\cite{QiangCWW17}, MRF model defines the binary potential (weight of undirected edge) for each edge $(z_{w_{i}}, z_{w_{j}})$ of undirected graph $G_d$ as  $\exp({\mathbbm{1}_{z_{w_{i}} = z_{w_{j}}}})$, where $\mathbbm{1}$ is the indicator function. $P_d$ is the set of edges and $|P_d|$ is the total number of edges in undirected graph $G_d$ for $d$th document. $\eta$ is an user-specified parameter that controls the effects of MRF Regularization into our model. If $\eta = 0$, then we do not consider the effect of MRF into our model.

We  employ Gibbs sampling to estimate the posterior distribution by sampling the variables of interest $z_t$ and $l_t$ here, for word $w_t$ from the distribution over the variables, given the current values of all  other variables and data. We now compute the joint probability distribution in Eq.~\ref{jointpWTSnew}.

\begin{equation} \label{probz}
\begin{split}
p(\mathbf{z})= {\left( \frac{\Gamma(\sum\limits_{j=1}^{T} \alpha_j)}{\prod\limits_{j=1}^{T}\Gamma(\alpha_j)} \right)}^{D}  \cdot \prod\limits_{d} \frac{\prod\limits_{j}\Gamma(N_{d,j} + \alpha_{j})}{\Gamma(N_{d} + \sum\limits_{j=1}^{T} \alpha_{j})} \cdot \\ 
 \exp({\eta \frac{\sum_{(z_{w_{a}},z_{w_{b}}) \in P_{d}}\sum_{j}{\mathbbm{1}_{z_{w_{a}}=z_{w_{b}}}}}{|P_{d}|}})
\end{split}
\end{equation} 

\begin{multline}
\label{computedJointpWTS}
\small
p(z_t = j, l_t = k \,|\, w_t,\mathbf{z}^{-{\bf t}},  \mathbf{l}^{-{\bf t}}, \pmb{\alpha}, \beta, \pmb{\gamma}) \propto  \\ 
 \frac{N_{j,k,w_t}^{-t} + \beta}{N_{j,k}^{-t} + V \beta} \cdot \frac{N_{d,j,k}^{-t} + \gamma_{d,k}}{N_{d,j}^{-t} + \sum\limits_{k}\gamma_{d,k}} \cdot \frac{N_{d,j}^{-t} + \alpha_j}{N_{d}^{-t} + \sum\limits_{j} \alpha_j} \cdot \\
 \exp(\eta \frac{\sum_{i \in N_{d_{w_{t}}}} \sum{\mathbbm{1}_{z_{i}=j}}}{|N_{d_{w_{t}}}|})
 \end{multline}
Above $N_{d_{w_{t}}}$ denotes the words appearing in the document $d$ that are labeled to be similar to word $w_t$ based on the embedding. Similarly, $|N_{d_{w_{t}}}|$ is the total number of such words. 
\\
We obtain samples from the Markov chain which are then used to approximate the per-corpus topic-sentiment word distribution:
\begin{equation}   \label{Mvarpsinew}
    \varphi_{j,k,i} =\frac{N_{j,k,i}+\beta}{N_{j,k}+V\beta}
    \end{equation} 
The per-document topic specific sentiment distribution is approximately computed as, 
\begin{equation} \label{Mpi}
\pi_{d,j,k} = \frac{N_{d,j,k}+\gamma_{d,k}}{N_{d,j}+ \sum\limits_{k} \gamma_{d,k}}
\end{equation}
Finally, we approximate per-document topic distribution as,
\begin{equation} \label{Mtheta}\small
\theta_{d,j} = \frac{N_{d,j} + \alpha_{j}}{N_{d} + \sum\limits_{j} \alpha_{j}}
\end{equation}

Algorithm~\ref{GibbsLJST} shows the pseudo-code for the Gibbs sampling procedure of ELJST.

\begin{algorithm}
{\small
\SetAlgoLined
\SetKwInOut{Input}{Input}
\SetKwInOut{Output}{Output}
\SetKwInOut{Assign}{Assign}
\SetKwInOut{Initialization}{Initialization}
\SetKwInOut{Iteration}{Iteration}
\SetKwInOut{Require}{Require}
\SetKwInOut{End}{End}
\SetKwInOut{Solution}{Solution}
\Input{${\pmb{\alpha}}, \beta, {\pmb{ \gamma}^{(d)}}$}
\Initialization{Initialize  matrix ${\mathbf \Theta}_{D \times T}$,  tensor ${\mathbf \Pi}_{D \times T \times S}$,  tensor $\Phi_{T \times S \times V}$;}
\For{i = 1 to {\em max} Gibbs sampling iterations}
     { \For{all documents $d \in \{1,2, \ldots, D\}$}
        { \For {all words $w_t,\, t \in \{1,2, \ldots, N_d\}$}
						{Exclude $w_t$ associated with topic $j$ and sentiment label $k$ and compute $N_{j,k,i}, N_j,k$ $N_{d,j,k}, N_{d,j}$, and $N_d$\; 	
						Sample a new topic-sentiment pair $\bar{z}$ and $\bar{k}$ using Eq.~\ref{computedJointpWTS}\;
		       Update variables $N_{j,k,i}, N_j,k$ $N_{d,j,k}, N_{d,j}$, and $N_d$  using the new topic label $\bar{z}$ and sentiment label $\bar{k}$\;
																	}
										}
										\eIf{number of iterations = {\em max} Gibbs sampling iterations}
										{Update $\Theta, \Pi$ and $\Phi$ with new sampling results given by Eqs~\ref{Mtheta},~\ref{Mpi} and~\ref{Mvarpsinew}  }{{\rm True}}
									}
}
 \caption{Gibbs sampling procedure for ELJST}
\label{GibbsLJST}
\end{algorithm}
\section{Experimental Setup}
In this section, we describe the setup for the experiments we have performed to demonstrate the effectiveness and robustness of our model over other baselines on 5 datasets.
\subsection{Datasets}
For experimental purposes we use three review datasets of 5-core from Amazon product data, {\it viz.}, Home, Kindle and Movies \footnote{\url{http://jmcauley.ucsd.edu/data/amazon/}}. We consider different kinds of datasets due to the fact that their writing styles vary across different categories and products. In order to show the efficacy of our model on short texts, we use an internal feedback data (IFD) gathered by our company on healthcare applications. IFD contains complaints and feedback collected from customers along with sentiment score provided by the customer on a scale of 1-5. \color{black}Further, we consider a dataset collected from Twitter, {\it viz.} Twitter Airline Data\footnote{\url{https://www.kaggle.com/crowdflower/twitter-airline-sentiment}} which can be considered as another corpus of short texts. Unlike other datasets, the Twitter dataset contains sentiment values on a scale of 0-2. \color{black}We use only the textual content of the datasets and associated sentiment values ignoring other attributes. For each of these datasets, we use standard text pre-processing techniques which involve converting all letters to lower cases, removing non-Latin characters and stop words, normalizing numeric  values into words and performing lemmatization on words. In order to make topics more interpretable and descriptive, we retain words with designated PoS labels such as Adjective, Adjective Comparative, Adjective Superlative, Noun (Singular/Plural), and Proper Noun (Singular/Plural). To construct the vocabulary for each corpus, we set the minimum document frequency to be 5 and the maximum to be 50\%. For further simplification, we choose 20,000 documents from each of the Amazon datasets using stratified sampling on the sentiment. As a standard practice, we use 80\% of our dataset for training and rest 20\% as test data. We produce the statistics of the datasets in Table~\ref{tab:dataset-statistics}. 

\begin{table*}[ht]
\caption{Statistics of the datasets after pre-processing.}
  \centering
  \scalebox{1}{
     \begin{tabular}{|l|r|r|r|r|r}
     \hline
    {\bf Dataset} & {\bf Corpus size} & {\bf Mean doc. length}  &  {\bf Mean sentiment} & {\bf Number of tokens}\\
    \hline
    Home  & 20,000 & 45  & 4.34 & 3,998\\
    Kindle  & 20,000 & 47  & 4.37 & 4,935\\
    Movies  & 20,000 & 52  & 4.17 & 5,863\\
    IFD  & 7,672 & 14  & 3.64 & 1,007\\
    \color{black} Twitter & \color{black} 9,061 & \color{black} 8 & \color{black} 0.41 & \color{black} 7,455 \\
    \hline
    \end{tabular}}
    
    \label{tab:dataset-statistics}%
\end{table*}%

\subsection{Baseline Methods}
We compare the performance of our model with five baselines mentioned below:
\begin{itemize}
    \item[$\circ$] Dependency-Sentiment-LDA (dsLDA)~\cite{Li10}, RJST~\cite{LinHER12} and TSM~\cite{MeiSZ07}, which are traditional LDA based joint topic-sentiment models.
    \item[$\circ$] ETM~\cite{QiangCWW17}, a short text topic model with word embedding. As ETM does not have sentiment associated, to make results comparable, we use $T \times S$ number of topics.
    \item[$\circ$] WS-TSWE~\cite{kbs/FuSWCH18,ecml/FuSWCH16} a weakly supervised joint topic-sentiment model with word embedding.
\end{itemize}
\subsection{Hyper-parameter Settings}
For document-topic distribution, we  chose $\pmb{\alpha}$ as the asymmetric prior. For initialisation, we empirically chose $\pmb{\alpha} = 10/T$, where $T$ is the number of topics. Similar to RJST, we use symmetric $\beta = 0.01$. The Dirichlet parameter $\gamma$ is the asymmetric prior as described in Section~\ref{ELJST}. For initialisation, we use $\gamma = 10/(T \times S)$. Depending upon the document sentiment label, $\gamma$ is different for each document. Also for test set, as mentioned in Section~\ref{ELJST}, we use only symmetric $\gamma$. For all the methods, same values for $\pmb{\alpha}, \beta$ and $\gamma$ are used. As suggested in ~\cite{QiangCWW17}, we use $\eta=1$ for both ETM and ELJST. For WS-TSWE we use $\lambda=0.1$ and $\mu=0.01$. In all the methods, Gibbs sampling is run for $1000$ iterations. The results reported in the paper are averaged over 5 runs.

\subsection{MRF Creation in ELJST} \label{expResult}
We construct a Markov Random Field by connecting semantically similar words with edges for each document. Words in a document are represented as vectors using a suitable word embedding. Recall from Section~\ref{proposedModel} that two words are semantically similar if distance between two word vectors using an appropriate distance metric is less than a threshold value ($\varepsilon$). \footnote{threshold value $\varepsilon$ is different from the perturbation value $\epsilon$ described in section \ref{ELJST}.} For representing words we use static word embedding - Word2Vec \footnote{\url{https://radimrehurek.com/gensim/models/word2vec.html}}, and sub-word level embedding - fastText \footnote{\url{https://fasttext.cc/}}. For base Word2Vec (without fine-tuning) we use 300-dimensional word embeddings trained on Google news data \footnote{\url{https://code.google.com/archive/p/word2vec/}}. In fine-tuning, we use 300-dimensional embeddings learned on each of the datasets separately. Word2Vec fine-tuning is done using Gensim with default parameter configuration for 20 epochs. Similarly, for fastText we use 300-dimensional embeddings trained on Common Crawl dataset \footnote{\url{https://dl.fbaipublicfiles.com/fasttext/vectors-english/crawl-300d-2M-subword.zip}}. In fine-tuning, we use 300-dimensional word embeddings learned on each dataset separately. fastText is fine-tuned using fastText library \footnote{\url{https://fasttext.cc/docs/en/unsupervised-tutorial.html}} with skip-gram for 20 epochs with a learning rate of $0.5$.
For unknown vocabulary words we use 300D vectors randomly sampled from glorot-uniform distribution. 

\color{black} We further use the BERT base model\footnote{\url{https://huggingface.co/transformers/model_doc/bert.html}} fine-tuned on our labelled dataset in the downstream classification task and extract the 768-dimensional vector representation for each word token. For BERT fine-tuning, we use Huggingface's \textit{BertForSequenceClassification} wrapper\footnote{\url{https://huggingface.co/transformers/model_doc/bert.html\#bertforsequenceclassification}} for sentiment classification task. We use the original pretrained BERT wordpiece tokenizer to tokenize our dataset.\footnote{\url{https://huggingface.co/transformers/model_doc/bert.html\#berttokenizer}} The classification model is trained on each of training datasets. We use Adam optimizer with a learning rate of $5e-5$ for $20$ epochs, with a early stopping of $5$ rounds on the validation dataset. We extract the 768-dimensional word token embeddings and multi-headed self-attention weights between each token pair from the base BERT model of the fine-tuned classification model. 
\color{black}
For all the embedding models, cosine similarity is used to measure the similarity between two word vectors. We vary the threshold between 0.3 and 0.9 to observe how the model changes with respect to loosely or densely connected Markov Fields. Best results are observed at $\varepsilon = 0.3$ for most of the embedding models. 

\begin{figure*}[ht]
\centerline{\includegraphics[scale=0.131]{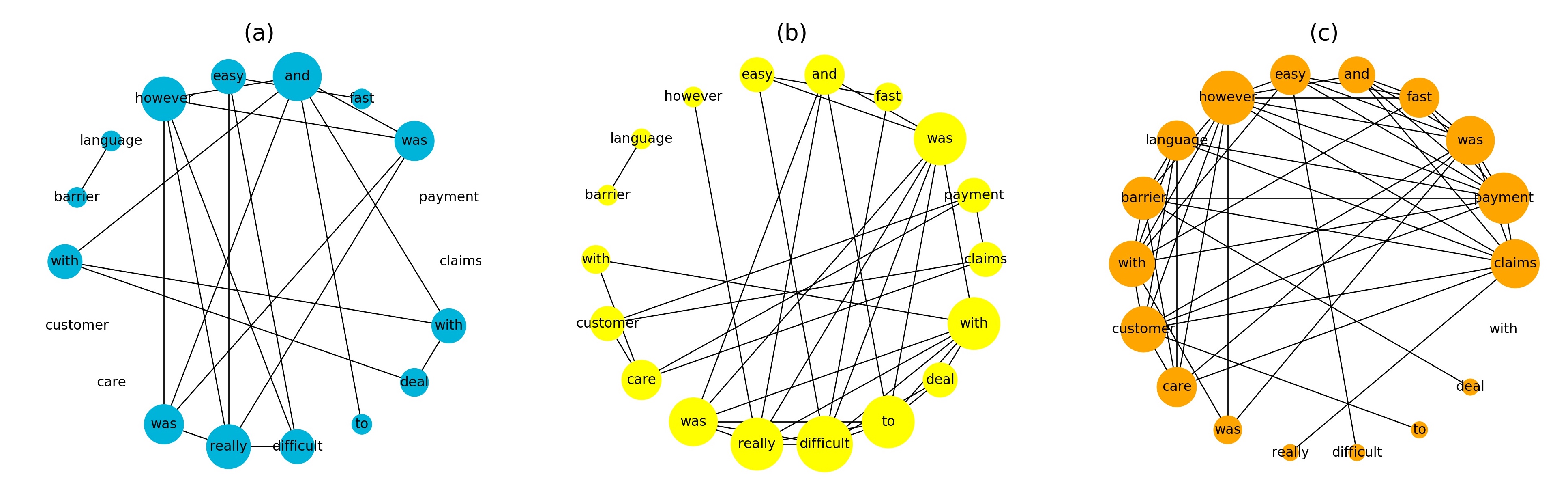}}
\caption{MRF creation for a sample document using different methods: (a) fastText (pre-trained) (b) BERT embedding (c) BERT attention. For fastText we use a threshold of $\varepsilon = 0.3$ and for BERT $\varepsilon = 0.9$. For BERT attention we choose only the token with highest attention value for each word.}
\label{textgraph}
\vspace{-0.35cm}
\end{figure*}

BERT-based model consists of 12 layers with 12 self-attention heads each. The attention heads operate in parallel and help the model capture wider range of relationships across words. We first considered all 12 heads from only the last layer. For each of the 12 attention heads, we pick the token with highest attention weight for each word; 
using this approach we observed that the learning at each attention head is different. Hence, we combine all the 12 heads by taking average to construct the undirected graph for each document. Two words within a document are connected by an edge, if and only if, they attend each other under any attention head. \color{black}Therefore, mathematically, we construct $e_{ij} = (w_i, w_j)$ between words $w_i$ and $w_j$ from any document $d$, if and only if
\begin{multline}
    j = \argminA_k \{ Attention^{head_{n}}(w_i,w_k); \\ \text{for some } n \in \{1,2,\ldots,12\}\}
\end{multline}
\begin{multline}
    i = \argminA_k \{ Attention^{head_{n}}(w_j,w_k); \\ \text{for some } n \in \{1,2,\ldots,12\}\}
\end{multline}
\color{black}
Note that a few edges from attention could form self loops which will not be considered here. 

Table~\ref{tab:edge-statistics} shows statistics of different embedding representations. The mean and average number of edges with a threshold 0.9 is quite low for almost all the models. Threshold 0.3 provides more edges for Word2Vec and fastText. fastText when fine tuned, generates word vectors that are more domain specific and in turn carries more information when computing similarities across words. BERT produce word vectors with high contextual information due to which the similarities are much closer and the threshold is chosen empirically. BERT generates too many edges for few documents making the maximum number of edges go much higher. BERT attention provides appropriate amount of edges where the mean number of edges is not too low and the maximum number of edges is not too high. BERT attention is observed to preserve local as well as global contexts much better than other variations.

\begin{table*}[ht]
\caption{Statistics of constructed number of edges (for each document) for different types of embedding}
\centering
{\scriptsize 
\begin{tabular}{|l|l|l|l|l|l|l|l|l|l|l|l|l|}
\hline
\multicolumn{1}{|c|}{\multirow{2}{*}{\textbf{Embedding}}} & \multicolumn{1}{c|}{\multirow{2}{*}{\textbf{Tuned}}} & \multicolumn{1}{c|}{\multirow{2}{*}{\textbf{Threshold $\varepsilon$}}} & \multicolumn{2}{c|}{\textbf{Kindle}} & \multicolumn{2}{c|}{\textbf{Movies}} & \multicolumn{2}{c|}{\textbf{Home}} & \multicolumn{2}{c|}{\textbf{IFD}} & \multicolumn{2}{c|}{\color{black} \textbf{Twitter}}\\ \cline{4-13} 
\multicolumn{1}{|c|}{} & \multicolumn{1}{c|}{} & \multicolumn{1}{c|}{} & \textbf{Mean} & \textbf{Max.} & \textbf{Mean} & \textbf{Max.} & \textbf{Mean} & \textbf{Max.} & \textbf{Mean} & \textbf{Max.} & \color{black} \textbf{Mean} & \color{black} \textbf{Max.} \\ \hline
Word2Vec & No & 0.3 & 25 & 258 & 24 & 209 & 23 & 204 & 7 & 31 & \color{black} 5 & \color{black} 24\\ \hline
Word2Vec & Yes & 0.3 & 28 & 312 & 31 & 239 & 29 & 276 & 9 & 35 & \color{black} 6 & \color{black} 27\\ \hline
Word2Vec & No & 0.9 & 3 & 19 & 3 & 18 & 3 & 19 & 3 & 7 & \color{black} 2 & \color{black} 5\\ \hline
Word2Vec & Yes & 0.9 & 4 & 21 & 5 & 20 & 5 & 27 & 5 & 10 & \color{black} 2 & \color{black} 6\\ \hline
fastText & No & 0.3 & 77 & 595 & 84 & 595 & 86 & 712 & 11 & 47 & \color{black} 6 & \color{black} 27\\ \hline
fastText & Yes & 0.3 & 83 & 617 & 84 & 601 & 87 & 719 & 13 & 53 & \color{black} 7 & \color{black} 29\\ \hline
fastText & No & 0.9 & 2 & 16 & 2 & 32 & 2 & 19 & 2 & 13 & \color{black} 2 & \color{black} 12\\ \hline
fastText & Yes & 0.9 & 2 & 16 & 2 & 33 & 3 & 21 & 2 & 14 & \color{black} 2 & \color{black} 15\\ \hline
BERT & Yes & 0.9 & 41 & 820 & 48 & 861 & 37 & 816 & 15 & 78 & \color{black} 11 & \color{black} 43\\ \hline
BERT attention & Yes & NA & 71 & 284 & 76 & 291 & 82 & 327 & 8 & 47 & \color{black} 10 & \color{black} 39\\ \hline
\end{tabular}
\vspace*{0.1cm}

\label{tab:edge-statistics}
}
\end{table*}

In Figure~\ref{textgraph} we show how different embedding methods help extracting different levels of knowledge from texts. A naive version of fastText embedding fails to capture semantic similarities between domain specific words - `customer', `care', 'language' and 'barrier', which is realised by other methods. Using BERT attentions we can capture the long term dependencies \footnote{one word appearing not in near neighbourhood of another word} between words `claims', `payment' and `easy' and similarly between `customer', `language' and `barrier'.


\section{Experimental Results}
We evaluate our results in a two-pronged manner, qualitative and quantitative.
\subsection{Quantitative Evaluation}
Our quantitative evaluation is based on measuring the (1) quality of topic sentiment model, (2) quality of topical representation of documents, and (3) quality of document modeling.

To measure the quality of extracted topic sentiment model we use the coherence metric, \textbf{Topic-Sentiment Coherence Score}~\cite{dieng2019topic} (TSCS),  which is defined as the average pointwise mutual information of word pairs under each topic-sentiment pair. 
The larger the TSCS value is, the tighter the word pairs are, which in turn makes topics more coherent and interpretable.

To measure the quality of topical representation we use \textbf{Diversity score}~\cite{dieng2019topic} which helps  understand the uniqueness of words generated per topic. 
A Diversity score close to 0 indicates redundant topics, whereas diversity close to 1 reflects more varied topics. 

As topic modeling is closely related to document clustering we use another topical representation quality metric \textbf{H-score}~\cite{Yan+13} based on Jensen-Leibler divergence between two documents. A low H-score implies that the average inter-cluster distance is larger than the average intra-cluster distance which results in tightly coupled clusters, hence the documents that share  similar topic distribution are close to each other.

To evaluate the generative behaviour of our model we compute the \textbf{perplexity}~\cite{BleiNJ03} of the test set. The lower the perplexity, the better is the generative performance of the model.

\begin{table}[ht]
\caption{ Performance of ELJST (with BERT attention) against the baselines. Best performance for all models are observed for Amazon and Twitter datasets at $T=5$ and for IFD at $T=10$ (where $T$ = no of topics).}
\hspace*{-0.7cm}
\flushright
{\scriptsize 
    \scalebox{0.85}{
    \begin{tabular}{l|l|ccc|l}
    \hline
     \multirow{2}{*}{\textbf{Data}} & \multirow{2}{*}{\textbf{Model}} & \multicolumn{3}{c|}{\bf Topic} & \multirow{2}{*}{\textbf{Perplexity}} \\ \cline{3-5} 
     &  &  \textbf{H-Score} & \textbf{TSCS} &  \textbf{Diversity}  &  \\
    \hline
    \multirow{6}{*}{Home} 
    & dsLDA & 0.362 & 0.174 & 0.410 & 5531.9 \\ 
    & ETM & 0.193 & 0.192 & \textbf{0.770} & 5717.4 \\ 
    & RJST & 0.360 & 0.131 & 0.620 & 5408.0 \\ 
    & TSM & 0.445 & 0.145 & 0.540 & 5966.5 \\ 
    & WS-TSWE & 0.253 & 0.203 & 0.710 & 5102.3 \\ 
    & \color{black} ELJST ($\eta = 0$) & \color{black} 0.329 & \color{black} 0.141 & \color{black} 0.600 & \color{black} 5201.7 \\
    & \textbf{ELJST} & \textbf{0.118} & \textbf{0.214} & 0.740 & \textbf{4957.2} \\ 
    \hline
    \multirow{6}{*}{Kindle}
    & dsLDA & 0.482 & 0.067 & 0.220 & 7643.2 \\ 
    & ETM & 0.200 & 0.182 & 0.650 & 6984.0 \\ 
    & RJST & 0.387 & 0.114 & 0.600 & 7967.3 \\ 
    & TSM & 0.477 & 0.134 & 0.560 & 7966.5 \\ 
    & WS-TSWE & 0.176 & 0.180 & 0.630 & 6766.4 \\ 
    & \color{black} ELJST ($\eta = 0$) & \color{black} 0.201 & \color{black} 0.097 & \color{black} 0.450 & \color{black} 7014.5 \\
    & \textbf{ELJST} & \textbf{0.113} & \textbf{0.196} & \textbf{0.710} & \textbf{6513.3} \\
    \hline
    \multirow{6}{*}{Movies} 
    & dsLDA & 0.488 & 0.166 & 0.380 & 5552.1 \\ 
    & ETM & 0.178 & 0.187 & 0.720 & 4467.6 \\ 
    & RJST & 0.367 & 0.090 & 0.630 & 5842.3 \\ 
    & TSM & 0.462 & 0.125 & 0.480 & 5991.7 \\ 
    & WS-TSWE & 0.445 & 0.194 & 0.690 & 4008.0 \\ 
    & \color{black} ELJST ($\eta = 0$) & \color{black} 0.337 & \color{black} 0.112 & \color{black} 0.710 & \color{black} 4590.1 \\
    & \textbf{ELJST} & \textbf{0.124} & \textbf{0.227} & \textbf{0.750} & \textbf{3834.7} \\
    \hline
    \multirow{6}{*}{IFD} 
    & dsLDA & 0.613 & 0.052 & 0.680 & 817.25 \\ 
    & ETM & 0.431 & 0.117 & 0.730 & 701.03 \\ 
    & RJST & 0.558 & 0.079 & 0.690 & 830.11 \\ 
    & TSM & 0.542 & 0.080 & 0.650 & 832.55 \\ 
    & WS-TSWE & 0.408 & 0.102 & 0.740 & 692.67 \\ 
    & \color{black} ELJST ($\eta = 0$) & \color{black} 0.529 & \color{black} 0.067 & \color{black} 0.630 & \color{black} 798.06 \\
    & \textbf{ELJST} & \textbf{0.301} & \textbf{0.126} & \textbf{0.740} & \textbf{681.09} \\
    \hline
    \multirow{6}{*}{\color{black} Twitter} 
    & \color{black} dsLDA & \color{black} 0.511 & \color{black} 0.057 & \color{black} 0.288 & \color{black} 1457.2 \\ 
    & \color{black} ETM &  \color{black} 0.157 & \color{black} 0.146 & \color{black} 0.300 & \color{black} 2208.4 \\ 
    & \color{black} RJST & \color{black} 0.498 & \color{black} 0.198 & \color{black} 0.336 & \color{black} 1434.3 \\ 
    & \color{black} TSM & \color{black} 0.492 & \color{black} 0.112 & \color{black} 0.264 & \color{black} 2033.3 \\ 
    & \color{black} WS-TSWE & \color{black} 0.224 & \color{black} 0.186 & \color{black} 0.144 & \color{black} 1012.8 \\ 
    & \color{black} ELJST ($\eta = 0$) & \color{black} 0.082 & \color{black} 0.173 & \color{black} 0.350 & \color{black} 279.63 \\
    & \color{black} \textbf{ELJST} & \color{black} \textbf{0.078} & \color{black} \textbf{0.201} & \color{black} \textbf{0.440} & \color{black} \textbf{280.75} \\
    \hline
    \end{tabular}}%
        \vspace*{0.1cm}
    	
	  \label{tab:allresults}%
	  }
	\hspace*{-3cm}
\end{table}%

Table~\ref{tab:allresults} shows the comparison of ELJST to the baseline methods on all the datasets. Among all embedding configurations, the best performance for ELJST is observed under BERT attention settings. For ETM and WS-TSWE however, the best results are observed with fastText fine-tuned embeddings. It is easy to see that ELJST consistently outperforms other baseline methods under all the evaluation metrics. JST based models such as dsLDA, RJST and TSM behave similarly as they are built on similar generative structure. 
On the other hand, both ETM and WS-TSWE perform much better in terms of topic quality, as they incorporate contextual information into the models through word embedding. The topic-sentiment pairs identified by ELJST are at least 8\% more coherent than the ones extracted by WS-TSWE and ETM models. On the other hand, we observe relatively low variability in the topic diversity, although ELJST demonstrates the highest diversity among all the models. 
In document clustering task, ELJST demonstrates a drastic improvement of over 20\% over other baselines. \color{black}Even on shorter texts in IFD and Twitter datasets, ELJST observes more than 30\% improvement in the document clustering and more than 1.5\% improvement in the topic coherence. In order to show the performance gain due to the utilization of labelled data, we also compare our method with the version with no MRF by setting $\eta = 0$. Table~\ref{tab:allresults} shows that ELJST with MRF always outperforms ELJST with $\eta = 0$. Even in most of the cases, the unsupervised baselines outperform no MRF version of ELJST. This shows the contribution of embeddings into our model. Hence, the superiority of our model is not just due to external labels, rather, due to underlying generative model structure and the ability to use the semantic information through different embeddings.
\color{black}
\paragraph{Performance of ELJST under different parameter settings:} \color{black} In Tables~\ref{tab:eljstcomparison} and ~\ref{tab:eljstcomparisonTwit}, we further explain the performance of different ELJST configurations on IFD and Twitter datasets. \color{black}We observe similar behavior in other datasets as well. ELJST modeled with $\eta = 0$ does not use the MRF regularizer, this model shows very little improvement over the JST models. Gradual improvement is observed when we add word level or sub-word level embedding. Further, fine-tuning of embedding models on individual datasets leads drastic improvement. As shown in Figure~\ref{textgraph}, pre-trained embedding fail to capture relationship between domain specific words and their polarities. Further slight modification helps retaining the original polarity as well as understanding the connection with domain specific keywords. Of all configurations, we find the BERT attention model captures semantically the most meaningful relationships.
Also it can preserve local properties (linkage between consecutive words) as well as global properties (long distance relationships), which is essential for coherent topic modelling. 

\begin{figure}[ht]
\centerline{\includegraphics[scale=0.25]{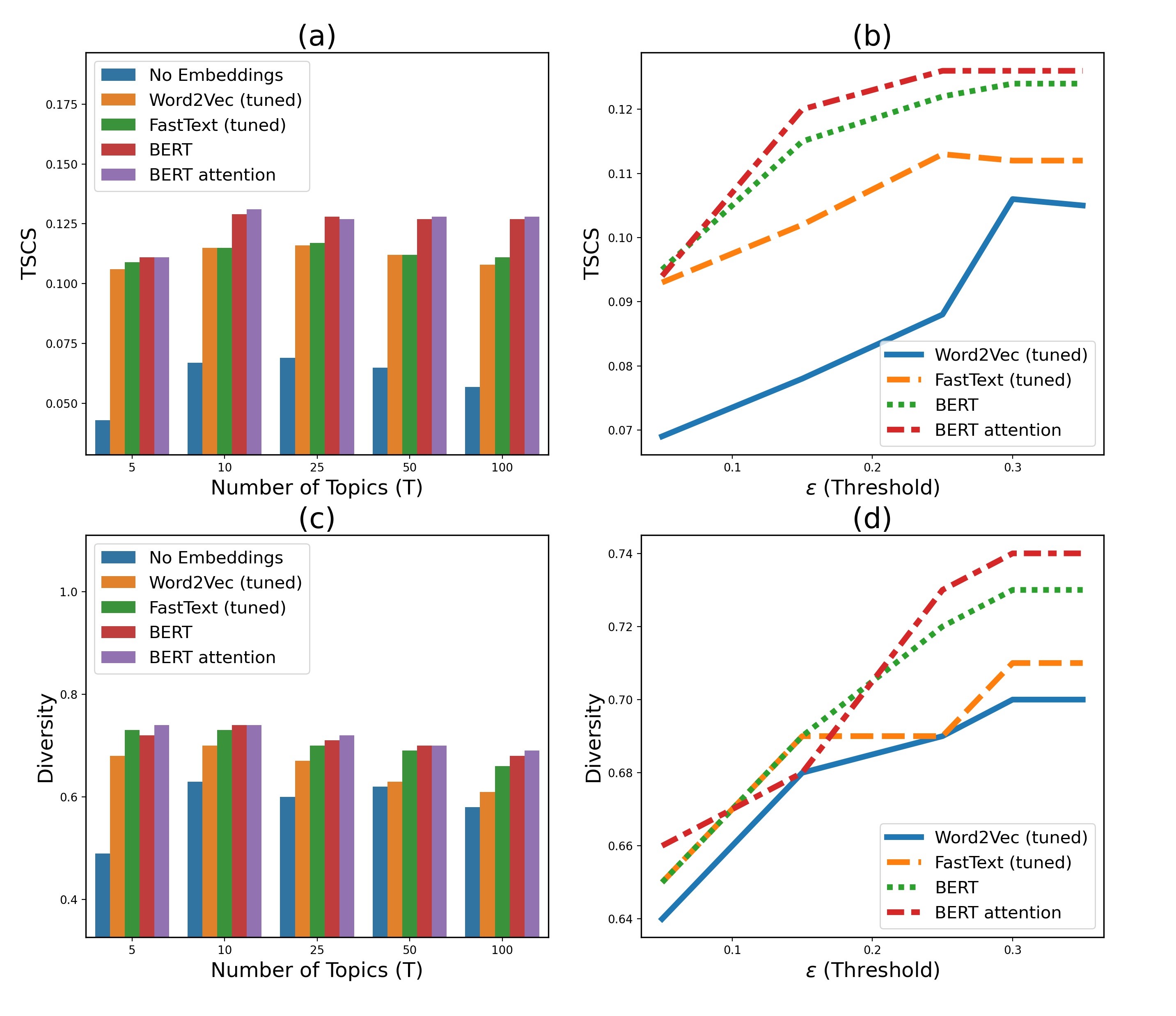}}
\hspace*{-.9cm}
\caption{Performance of the different settings of ELJST model with varying $T$ (keeping $\varepsilon = 0.3$) and $\varepsilon$ (keeping $T = 10$) on IFD dataset for topic coherence evaluation (a,c) and topic diversity (b,d).} 
\label{fig:accuracy}
\vspace{-2.05mm}
\end{figure}

\begin{figure}[ht]
\centerline{\includegraphics[scale=0.25]{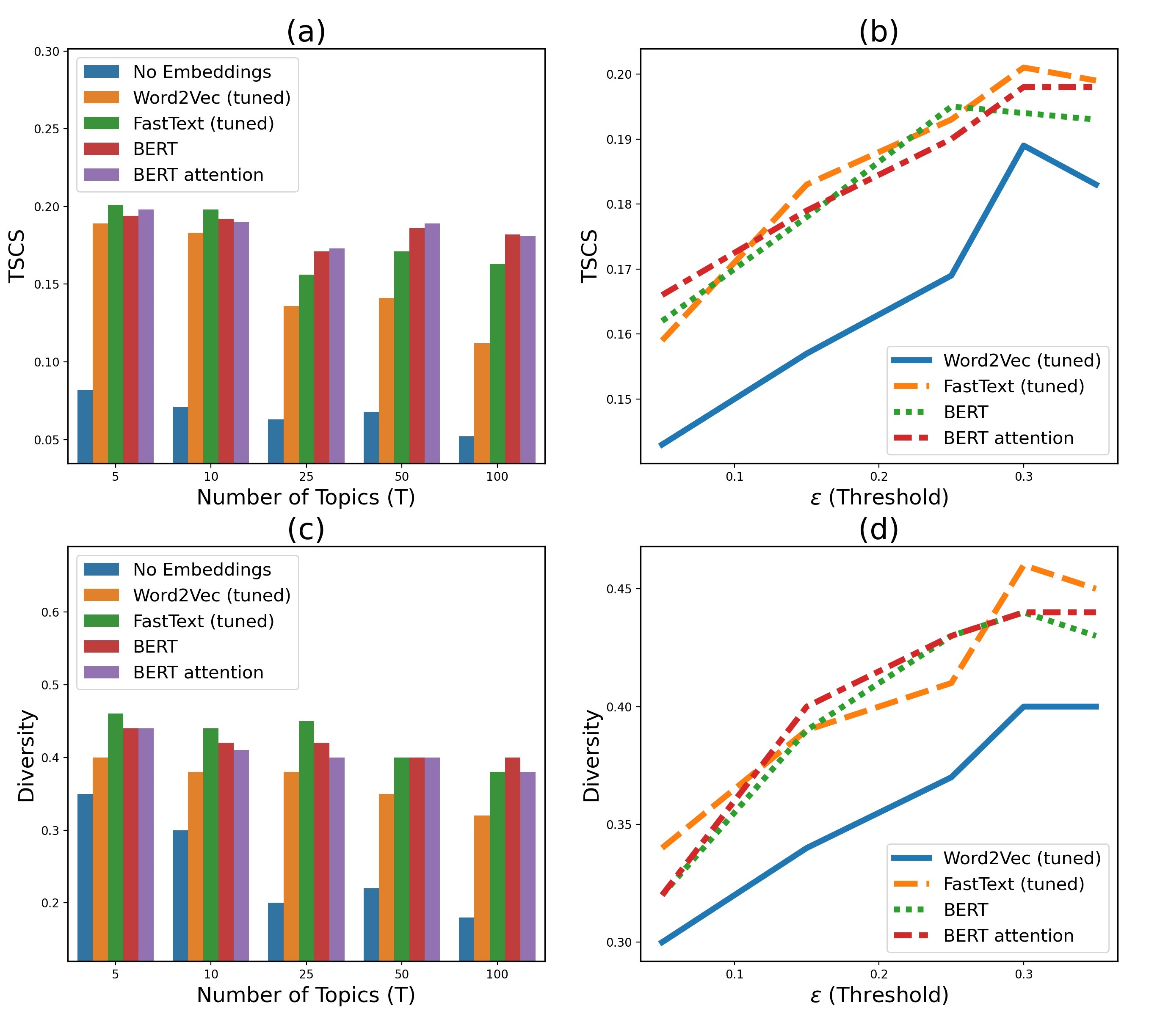}}
\hspace*{-.9cm}
\caption{Performance of the different settings of ELJST model with varying $T$ (keeping $\varepsilon = 0.3$) and $\varepsilon$ (keeping $T = 5$) on Twitter dataset for topic coherence evaluation (a,c) and topic diversity (b,d).} 
\label{fig:accuracytwitter}
\vspace{-2.05mm}
\end{figure}

\color{black}In Figures~\ref{fig:accuracy} and  ~\ref{fig:accuracytwitter}, we further show the performances of different ELJST configurations under different parameter settings. \color{black}In ELJST we take $S$ (number of sentiment labels) to be the same as the number of unique classes in the labelled data. Therefore, we only vary the number of topics ($T$) and the threshold parameter ($\varepsilon$). As described in Table~\ref{tab:edge-statistics}, we observe that increasing $\varepsilon$ makes the undirected graph sparse, resulting in the reduction in the the ability of MRF regularizer. We observe the downward trend in TSCS and diversity score with increasing $\varepsilon$. On the other hand, increasing $T$ can lead to detected topics being more similar to each other. However, there is a trade off between coherence and diversity when we increase the number of topics. 

\begin{table}[ht]
\caption{{\small Comparison of ELJST variants on IFD.}}
\centering
\small
\scalebox{0.8}{
\begin{tabular}{|l|l|l|l|l|l|}
\hline
\textbf{Embedding} & \textbf{Tuned} & \textbf{HScore} & \textbf{TSCS} & \textbf{Div.} & \textbf{Perpl.} \\ \hline
None ($\eta = 0$) & - & 0.529 & 0.067 & 0.630 & 798.06 \\ \hline
Word2Vec & No & 0.411 & 0.106 & 0.700 & 701.24 \\ \hline
Word2Vec & Yes & 0.401 & 0.115 & 0.710 & 698.17 \\ \hline
fastText & No & 0.402 & 0.113 & 0.710 & 693.28 \\ \hline
fastText & Yes & 0.341 & 0.124 & 0.730 & 690.44 \\ \hline
BERT & Yes & 0.312 & 0.125 & \textbf{0.740} & 685.75 \\ \hline
BERT Attention & Yes & \textbf{0.301} & \textbf{0.126} & \textbf{0.740} & \textbf{681.09} \\ \hline
\end{tabular}}

\label{tab:eljstcomparison}%
\end{table}

\begin{table}[ht]
\color{black}
\caption{{\small Comparison of ELJST variants on Twitter dataset.}}
\centering
\small
\scalebox{0.8}{
\begin{tabular}{|l|l|l|l|l|l|}
\hline
\textbf{Embedding} & \textbf{Tuned} & \textbf{HScore} & \textbf{TSCS} & \textbf{Div.} & \textbf{Perpl.} \\ \hline
None ($\eta = 0$) & - & 0.082 &	0.173 & 0.350 & 279.63\\ \hline
Word2Vec & No & 0.087 & 0.189 & 0.380 & 277.99 \\ \hline
Word2Vec & Yes & 0.078 & 0.180 & 0.400 & 263.03 \\ \hline
fastText & No & \textbf{0.074} & 0.190 & 0.440 & 277.25 \\ \hline
fastText & Yes & 0.082 & \textbf{0.201} & \textbf{0.460} & \textbf{242.49} \\ \hline
BERT & Yes & 0.089 & 0.194 & 0.440 & 286.17 \\ \hline
BERT Attention & Yes & 0.078 & 0.198 & 0.440 & 280.75 \\ \hline
\end{tabular}}

\label{tab:eljstcomparisonTwit}%
\end{table}

\subsection{Qualitative Evaluation}
\begin{table*}[ht]

\centering
\begin{tabular}{cccccc}
\hline
\multicolumn{2}{c}{\textbf{Customer service}} & \multicolumn{2}{c}{\textbf{Rx order}} & \multicolumn{2}{c}{\textbf{Claims}}  \\ \hline \hline
positive & negative & positive & negative & positive & negative \\ \hline
\multicolumn{6}{c}{\textbf{ELJST}}\\ \hline
professional & dead & medicine & afford* & policy & copay \\
know* & information & \#\#fill* & return & coverage & payment\\
customer & hang & delivery & expensive & payment & rebut\\
efficient & unavailable & fast & unclear & reimburs* & expensive\\
language & rude & free & late & authorization & denied\\
\hline
\multicolumn{6}{c}{\textbf{WS-TSWE}}\\ \hline
customer & rude & medicine & expensive & authorization & payment \\
callback & difficult & doctor & return & prior & surgery\\
excellent & horrible & prescription & rx & network & waiting\\
phone & hang & clear & deliver & surgery & approval\\
prompt & waiting & fast & late & great & reject\\
\hline
\multicolumn{6}{c}{\textbf{RJST}}\\ \hline
customer & rude & delivery & late & policy & network \\
great & drop & medicine & return & claim & cost\\
excellent & hang & fast & cost & hospital & expensive\\
timely & horrible & great & expensive & doctor & charges\\
conversation & pathetic & perfect & medicine & helpful & frustrating\\
\hline
\end{tabular}
\caption{Top 5 words under positive and negative sentiment levels for 3 topics from IFD (* denotes wordpiece).}
\label{fig:topwords}

\end{table*}

In qualitative evaluate, we observe the top words detected by ELJST (with BERT attention weights) under different topic-sentiment pair. In Table~\ref{fig:topwords} we show the topic-sentiment of our model compared to two other baselines for IFD. We show top words under positive and negative sentiment labels, where we assume rating $1$ or $2$ to be negative and $4$ or $5$ to be positive. Traditional topic models tend to pick up most frequently occurring words and word pairs under topics. Typically in e-commerce or retail domain, top frequent words are adjectives or names of products. Therefore, topics become colluded with same words, which do not show actionable insights. On the other hand, ELJST, due to the regularization factor, tend to assign high coherent word pairs under different topics. Further use of overall text sentiment helps it understand the difference between word pairs with different sentiment polarities. This leads to highly diverse set of topics for each of the sentiment classes. which lead to more coherent word pairs. As shown in Table~\ref{fig:topwords}, ELJST picks ``knowledgeable" and ``efficient" under positive sentiment for topic {\tt Customer service}, thus it is able to understand the context as well as the correct polarity given the context. Similarly, words ``return", ``expensive" and ``afford" are used as negative terms in the context of {\tt Rx order}  (medicine order). Both WS-TSWE and RJST use external word-sentiment lexicons, which allow them to detect ``great", ``excellent", ``clear" under positive sentiments and similarly ``expensive", ``late", ``horrible", ``difficult" under negative sentiments. However, domain specific keywords are not often understood by these models, due to lack of knowledge in the lexicon files. On the other hand, ELJST can understand the correct sentiment polarity even for domain specific words like ``knowledgeable", ``unavailable", ``free", ``reject" etc. Additionally, with the use of fine-tuned embeddings, ELJST can put different domain-specific contextually meaningful words under relevant topic-sentiment level. With this ELJST can extract highly human interpretable results. 
\section{Conclusion}
In this paper, we propose ELJST, a novel framework for joint extraction of sentiment and topics, particularly for short texts. Our proposed models are informed by the external sentiment labels which in turn, reinforce the extraction of better topics, and predict better sentiment scores. In ELJST model, we use MRF graph with word embedding representations, include attention models to compute the similarity between word in the graph. Interestingly these attention models, which have been used for the first time for this purpose,  help joint topic sentiment discovery achieve the best performance. \color{black}Although the use of labeled text data in the model restricts the applicability of ELJST in many applications, ELJST  can be used in various applications across different industries, particularly, in the e-commerce and service based companies where sentiment/ratings are automatically labeled by the end customers. ELJST is currently deployed in a healthcare application which is helping with VoC (Voice of Customer) analysis and NPS (Net Promoter Score) improvement initiatives. In these two applications, ELJST helps in extracting granular level information from survey and complaint texts shared by customers (along with the discrete rating value on a scale of 1-5) and helps in creating  value for their service and enhancing customer satisfaction. \color{black}

\newpage
\bibliography{anthology, aacl-ijcnlp2020}
\bibliographystyle{aaai}

\end{document}